\documentclass{llncs}

\usepackage[utf8x]{inputenc}
\usepackage{url}
\usepackage{hyperref}
\usepackage{amsfonts}
\usepackage{amsmath}
\usepackage{amssymb}
\usepackage{graphicx}
\usepackage{epsfig}
\usepackage{caption}
\usepackage{subcaption}

\newcommand{\cphydra}{CPhydra}

\newcommand{\satzilla}{SATzilla}

\makeatletter
\def\url@leostyle{%
  \@ifundefined{selectfont}{\def\UrlFont{\sf}}{\def\UrlFont{\small\ttfamily}}}
\makeatother
\begin{document}

\title{An Empirical Evaluation of Portfolios Approaches for solving CSPs}
\subtitle{(Long Version)}

\author{Roberto Amadini \and Maurizio Gabbrielli \and Jacopo Mauro}

\institute{
 Department of Computer Science and Engineering/Lab. Focus INRIA,\\ University of Bologna, Italy.\\
 \email{\{amadini, gabbri, jmauro\}@cs.unibo.it}
}

\maketitle

\begin{abstract}
Recent research in areas such as SAT solving and Integer Linear Programming 
has shown that the performances of a single arbitrarily efficient solver can 
be significantly outperformed by a portfolio of possibly slower on-average solvers.
We report an empirical evaluation and comparison of portfolio approaches applied to 
Constraint Satisfaction Problems (CSPs).
We compared models developed on top of off-the-shelf machine learning algorithms with respect to 
approaches used in the SAT field and adapted for CSPs, considering different portfolio sizes and using as evaluation metrics the number of solved pro\-blems 
and the time taken to solve them.
Results indicate that the best SAT approaches have top performances also in the CSP field and are slightly more competitive than simple models built on top of classification algorithms.
\end{abstract}

\section{Introduction}

The past decade has witnessed a significant increase in the number
of constraint solving  systems deployed for solving
constraint satisfaction problems (CSP).
It is well recognized within the field of constraint programming that
different solvers are better at solving different problem instances,
even within the same problem class~\cite{DBLP:journals/ai/GomesS01}.
It has also been shown in other areas, such as satisfiability
testing~\cite{DBLP:conf/cp/XuHHL07} and
integer linear programming~\cite{DBLP:conf/cp/Leyton-BrownNS02},
that the best on-average solver can be
out performed by a \emph{portfolio} of possibly slower on-average solvers.
This selection process is usually performed by using machine learning techniques
based on feature data extracted from the instances that need to be solved.
Thus in general a Portfolio Approach \cite{DBLP:journals/ai/GomesS01} is a methodology 
that exploits the significant variety in performance observed between different  algorithms  and 
combines them in a portfolio to create a globally better solver.

Portfolio approaches in particular have been extensively studied and used in the SAT solving field.
Since historically the boolean sa\-tis\-fia\-bi\-li\-ty testing is the prototypical and one of the simplest NP-complete problems it has attracted a lot of attention and, over the last years, it has seen a tremendous progress. Problems 
that seemed to be completely out of reach a decade ago can now be solved by using new algorithms, better heuristics, and refined implementation techniques. Starting from 2002, a competition was held annually to evaluate the performances of different solvers and a big set of real case, random generated, and handcrafted instances were defined in a standard language (Dimacs format). The large number of different solvers available, the presence of a standard input language, and a huge dataset of instances has supported and fostered the study of how different solvers can be exploited in order to be able to solve more instances in a faster way.

On the other hand, to the best of our knowledge in the CSP field there exists only one solver
that uses a portfolio approach, namely \cphydra \cite{cphydra}. This solver uses a rather small portfolio (consisting of only 3 solvers) and seems rather limited when compared to modern SAT portfolio approaches.

Given this situation, in this work we tried to investigate to what extent a portfolio approach can increase the performances of a CSP solver 
and which could be the best portfolio approaches, among the several existing, for CSPs. In a portfolio approach of course it is important 
the quality of the solvers included in the portfolio. However, as previously mentioned, it is also essential the technique which is used  
in the selection of the different solvers.
Hence, in order to perform our study, we considered 22 versions of 6 well known CSP solvers, namely AbsCon (2 versions), BPSolver, Choco (2 versions), Mistral, Sat4j (all these solvers participated to the International CSP Solver Competition)
and 15 different versions of Gecode.
Using these 22 solvers we implemented two classes of CSP portfolio solvers, building portfolios of up to 16 solvers: in the first class we used relatively simple, off-the-shelf machine learning classification algorithms in order to define solver selectors; in the second class we tried to adapt the best, evolute, and complex approaches of SAT solving to CSP.  A third portfolio solver that we considered was \cphydra, mentioned above. We then performed an empirical evaluation and comparison of these three different portfolio approaches. We hope that our results, described in the remaining of this paper, may lead to new insights, to a confirmation of the quality of some approaches and also to some empirical data supporting the creation of better and faster CSP solvers. 

It is worth noticing that adapting portfolios techniques from other fields is not trivial: for instance, since portfolio approaches usually exploit features extracted from the various instances of the problems, a good selection of the features may be the responsible of the quality and the performances of an approach.
Moreover, differently from the SAT world, in the CSP field there is no a standard language to express CSP instances, there are fewer solvers, and sometimes only few features and constraints are supported. To overcome these limitations we tried to collect a dataset of CSP instances  as extensive as possible and encoded them into XCSP \cite{DBLP:journals/corr/abs-0902-2362}, an XML-based language used to express constraints. We used this dataset to evaluate the performances  of the three different CSP portfolio approaches.  

The remaining of this paper is organized as follows. In Section \ref{prelim} we introduce some basic notions and we give an overview of the most successful portfolios approaches proposed in the literature. In Section \ref{DFS} we describe our dataset, what features were extracted from every instance and what solvers were used. In Section \ref{exp} and \ref{results} we present the experiments methodology and the results we obtained. Finally, section \ref{related} discusses some related work while section \ref{concl} contains some concluding remarks. All the code developed to conduct the experiments is available at \url{http://www.cs.unibo.it/~amadini/cpaior_2013.zip}.

\section{Preliminaries}
\label{prelim}
In this section, after introducing some basic general concepts, we describe \cphydra\ 
and the SAT specific portfolio approaches 
that we have adapted to CSP. 

\subsubsection{Background}
A \textit{Constraint Satisfaction Problem} (CSP) consists of a finite set of
va\-ria\-bles, each of which associated with a domain of possible values that the
variable could take, and a set of constraints that define the set
of allowed assignments of values to the
variables~\cite{DBLP:journals/ai/Mackworth77}.
Given a CSP the goal is normally to find a solution, that is an assignment to the
variables that satisfies all the constraints of the problem.

\textit{Machine Learning} (ML) is a broad field that uses concepts from computer science, mathematics,
statistics, information theory, complexity theory, biology and cognitive 
science~\cite{Mitchell97} to ``construct computer programs that automatically improve with experience''.
In this paper we are particularly interested in \emph{classification}, which 
is a well-known ML problem that consists of identifying to which of a set of 
categories (classes) a new observation belongs by means of appropriate 
classifiers. A classifier is therefore a function that maps a new \emph{instance} 
- characterized by one or more discrete or continuous features - to one of a finite 
number of classes~\cite{Mitchell97} on the basis of a training set of instances 
whose class is already known, trying to exploit such knowledge to properly 
classify each new instance. Our simplest models are built on top of the most common classifiers 
provided by the WEKA~\cite{Hall_theweka} tool, an open source software written in JAVA that 
implements a collection of ML algorithms for data mining tasks.
 
\subsubsection{CPhydra}
To our knowledge \cphydra~\cite{cphydra} is the only CSP solver which uses a portfolio approach. 
This solver uses a $k$-nearest neighbor algorithm in order to compute a schedule of the portfolio constituent solvers which maximizes the chances of solving an instance within a time-out of 1800 seconds. A weak point of \cphydra \ is that it is not scalable w.r.t. the number of the constituent solvers. This is due to the fact that finding an optimal schedule of the solvers is an NP-hard problem.
Nevertheless, using a small size portfolio, \cphydra~was able to win the 2008 International CSP Solver Competition.

\subsubsection{SAT Solver Selector (3S)}
3S~\cite{DBLP:conf/cp/KadiogluMSSS11} is
a SAT solver that conjugates a fixed-time static solver schedule with the dynamic selection 
of one long-running component solver.
Exploiting the fact that a lot of SAT instances are extremely easy for one solver and almost impossible to solve for the others, 3S first executes for 10\% of its time short runs of solvers. The schedule of solvers, obtained by solving an optimization problem similar to the one tackled by \cphydra, is computed offline (i.e. during the learning phase on training data). Then, at run time, if a given instance is not yet solved after the short runs a designated solver is executed for the remaining time. This solver is chosen among the ones that are able to solve the majority of the most $k$-similar instances in the training dataset.
3S solves the scalability issues of \cphydra\ because the schedule computation is done offline and it uses some techniques that speed up the search.
This allowed 3S to use a portfolio of $21$ solvers and be the best-performing dynamic portfolio at the International SAT Competition 2011.

%
%
\subsubsection{SATzilla}
\satzilla~\cite{DBLP:conf/cp/XuHHL07} is a SAT solver that relies on runtime prediction mo\-dels to 
select the solver that (hopefully) has the fastest running time on a given problem instance.
In the International SAT Competition 2009
, \satzilla\ won all three major tracks of the competition. 
More recently a new powerful version of \satzilla\ has been proposed~\cite{DBLP:conf/sat/XuHHL12}. Instead of using 
regression-based runtime predictions, the newer version uses a weighted random forest approach provided with an 
explicit cost-sensitive loss function punishing misclassifications in direct proportion to their 
impact on portfolio performance. This last version consistently outperforms the previous versions of \satzilla\ and the other competitors of the SAT Challenge 2012
in the Sequential Portfolio Track.

\subsubsection{ISAC}
In \cite{DBLP:conf/cpaior/MalitskyS12} the Instance-Specific Algorithm Configuration tool ISAC~\cite{DBLP:conf/ecai/KadiogluMST10} has been used as solver selector. Given a highly parametrized solver for a SAT instance, the aim of ISAC is to optimally tune the solver parameters on the basis of the given instance features. ISAC statically clusters every training instance by the $g$-means algorithm~\cite{DBLP:conf/nips/HamerlyE03} according 
to its normalized feature vector and then identifies the best tuning of pa\-ra\-me\-ters for the instances of each cluster employing the GGA algorithm~\cite{gga}. When a new instance needs to be classified, ISAC determines the cluster with the nearest center to the instance and selects the precomputed parameters for such cluster.
ISAC can be easily seen as a ge\-ne\-ra\-li\-za\-tion of an algorithm selector since it could be used to cluster the instances and when a new instance is encountered it selects the solver that solved the largest number of instances belonging to the nearest cluster.



\section{Solvers, Features and Dataset }
\label{DFS}
In this section we introduce the three main ingredients of our portfolios, that is:
the CSP solvers that we use; the features, extracted from the CSP instances, which 
are used in the machine learning algorithms; the dataset used to perform the tests.

\subsubsection{Solvers}
We decided to build our portfolios by using some of the solvers 
of the International CSP Solver Competition.
\footnote{The other possible choice would have been the MiniZinc Challenge.
We discarded this option because the MiniZinc Challenge involves less solvers and it targets also optimization problems.}
We were able to use 5 solvers of this competition, namely AbsCon (2 versions),
BPSolver, Choco (2 versions),
Mistral and Sat4j. Moreover, by using a specific plug-in described in 
\cite{DBLP:conf/cilc/MoraraMG11}, we were able to use also 15 different versions 
of the constraint solver Gecode (these different versions were obtained by tuning the search parameters 
and the variable selection criteria of the solver; the plug-in, that some of these authors developed, 
allowed Gecode to receive XCSP format in input). Thus we had the possibility of using, in our portfolio, up to 22 specific solvers which 
were all able to process CSP instances defined in the XCSP format \cite{DBLP:journals/corr/abs-0902-2362}.



\subsubsection{Features}
In order to train the classifiers, we extrapolated a set of 44 features from each XCSP instance. An extensive description of the features can be retrieved in \cite{KMMO11}. 
We used the 36 features of \cphydra~\cite{cphydra}
plus some features derived from the \emph{variable graph} and \emph{variable-constraint graph} 
of the XCSP instances.
Whilst the majority of these features are syntactical, some of them are computed by collecting 
data from short runs of the Mistral solver.
Among the syntactical features we can mention the number of variables, the number of 
constraints and global constraints, the number of constants, the size of the domains and the 
arity of the predicates. The dynamic features instead take into account the number of nodes 
explored and the number of propagations done by Mistral within a time limit of 2 seconds.
The time needed to compute these features is often negligible.\footnote{For the instances of the dataset the average feature computation time was 2.47 seconds with a standard deviation of 3.54 and a maximum of 93.1 seconds.}

\subsubsection{Dataset}
We tried to perform our experiments on a set of instances as rea\-lis\-tic and large as possible. 
Hence, we constructed a comprehensive dataset of CSPs based on the instances gathered 
from the 2008 International CSP Solver Competition\footnote{The last competition was held 
in 2009 and it did not introduce new instances in the dataset.} that are publicly available and already in a XCSP normalized format.
Moreover, we added to the dataset the instances from the MiniZinc suite benchmark. 
These instances written in FlatZinc~\cite{Nethercote07minizinc:towards} were first compiled to XCSP (by using a FlatZinc to XCSP converter provided by the MiniZinc suite)
and then normalized following the CSP competition conventions. Unfortunately, since FlatZinc is 
more expressive than XCSP not all the instances could be successfully converted. 

The final benchmark was built by considering 7163 CSP instances taken from the Constraint 
Competition, 2419 CSP instances obtained by the conversion of the MiniZinc instances and then discarding all the instances solved by Mistral during the first 
2 seconds computation of the dynamic features. We obtained a dataset containing 4547 instances (3554 from the Constraint Competition and 993 from MiniZinc).
For all the instances in the dataset we run all the 22 version of the solvers\footnote{We used Intel$^{\text{\textregistered}}$ 
Dual-Core 2.93GHz computers with 2 GB of RAM and Ubuntu operating system.} collecting their results and computation times with a time limit of 1800 seconds.
Among the dataset instances, $797$ could not be solved by any solver in our portfolio within the time cap.

\begin{figure}[ht]
\centering
\begin{subfigure}[b]{0.495\textwidth}
 \centering
 \includegraphics[width=\textwidth]{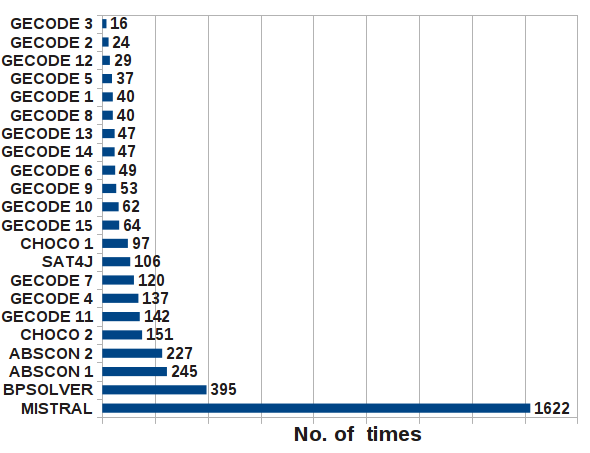}
 \caption{No. of times a solver is faster}
 \label{beat_all_others}
\end{subfigure}
\begin{subfigure}[b]{0.495\textwidth}
 \centering
 \includegraphics[width=\textwidth]{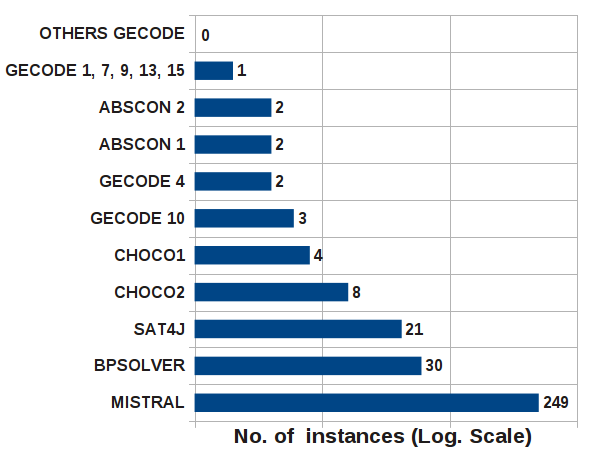}
 \caption{Marginal contributions}
 \label{one_and_only}
\end{subfigure}
\caption{Solver statistics}
\end{figure}
Figure \ref{beat_all_others} indicates the relative speed of the different solvers by showing,  
for each solver, the number of instances on which the con\-si\-de\-red solver is the fastest one. As it can be seen 
Mistral is by far the best solver,  since it is faster than the others for 1622 instances (36\% of the instances of the dataset).
In Figure \ref{one_and_only} following \cite{DBLP:conf/sat/XuHHL12} we show instead the marginal contributions of each solver, that is how many times a solver is 
able to solve instances that no other solver can solve. Even in this case Mistral is by far the best solver, almost one order of magnitude better than the second one.
It is worth noticing that there are also 8 versions of Gecode that do not give a marginal contribution.

\section{Methodology}
\label{exp}
In this section we present the methodology used to conduct the various experiments.
\subsubsection{Data Validation}
In order to evaluate and compare different portfolio approaches we tested every approach using a 5-repeated 5-fold cross-validation \cite{CrossVal}.
The dataset was randomly partitioned in 5 disjoint sets called folds. Each of these folds was treated in turn as the test set, considering the union of the 4 remaining folds as training data.
In order to avoid a possible overfitting problem (i.e. a portfolio approach that adapts too well on the training data rather than learning and exploiting the generalized pattern) the random gene\-ration of the folds was repeated 5 times, thus obtaining 25 sets of instances used to test the portfolio approaches. Every test set was therefore constituted by approximatively 909 instances
and the portfolio approach for a single fold was built by taking into account (approximatively) 3638 training instances.
For every instance of every test set we computed the solving strategy proposed by the portfolio approach and we simulated it by using a time cap of 1800 seconds, checking if the solving strategy was able to solve the instance and the time required.
To evaluate the performances of the portfolio approach we computed the ave\-ra\-ge solving time (AST) of the portfolio solver and the percentage of solved instances (PSI) for all the instances of the 25 test sets.

It is worth noticing that in order to evaluate the performance we simulated the execution of the solvers considering the solving times computed according to 
the description in the previous Section. Thus we implicitly assumed that all the solvers are deterministic and different run on  the same instance will produce the same results in the same time.
Moreover, in order to present a more realistic scenario, we have considered in  the simulation also the time taken to compute the instance features, 
even though usually this time is very short.

\subsubsection{Portfolios}
All the portfolio approaches were tested with portfolios of different sizes. Since we realized that some solvers had a very low marginal contribution
we considered portfolios consisting of up to a maximum of $16$ solvers.\footnote{The use of larger portfolios could have just reduce the best AST but not the PSI.} For every size $n = 2, \dots, 16$ the portfolio composition was computed by using a local search algorithm that maximized the number of instances solved by one of the solvers in the portfolio. Possible ties were broken by minimizing the average solving time for the instances of the dataset by the solvers in the portfolio.

\subsubsection{Off-the-shelf approaches}
For the approaches that used off-the-shelf machine learning classification algorithms we used a training set to train a classifier in order to select the best solver among those in the portfolio. For the instances that were not solved by any solver we added a new label \emph{no solver} that could be predicted. For every instance of the test set we simulated the execution of the solver selected by the model. In case the predicted solver was labeled \emph{no solver} or it finished unexpectedly\footnote{We experienced some solver failures due to bugs or unsupported constraint.} before the time cap the execution of a \emph{backup solver} was simulated for the remaining time.
To decide the best backup solver we exploited the Computational Social Choice theory~\cite{conf/sofsem/ChevaleyreELM07} mapping the selection problem into a voting scenario. We considered CSPs as voters who have to elect a representative among the 22 candidates solvers. Each CSP could express one or more preferences according to its favorite solver (i.e., the solver that solves it in less time). 
We simulated the elections using different positional scoring rules: Plurality (i.e. each CSP expresses at most one alternative), Approval (each CSP expresses a possibly empty set of favorite candidates), and Borda (a variant of Approval where votes are weighted). The election outcomes clearly sustained Mistral as the backup solver since it was the Condorcet winner, i.e. the candidate preferred by more voters when compared with every other candidate.

To train the models we used the WEKA tool \cite{Hall_theweka} which implements some of the most well known and widely used classification algorithms. In particular we used a $k$-nearest neighbors algorithm (IBk), decision trees based algorithms (RandomFo\-rest, J48, DecisionStump), bayesian networks (NaiveBayes), rule based algorithms (PART, OneR), support vector machines (SMO), and meta classifiers (AdaBoostM1, LogitBoost).\footnote{For more details related to these algorithms we defer to the documentation of \cite{Hall_theweka}.}
For all the classification algorithms we tried different parameters in order to increase their accuracy. This task was performed following the best practices when they were available or manually trying different parameters starting from the default ones of WEKA. For instance, for the support vector machine we used a Radial Basis Function kernel performing a grid search over the $C$ and $\gamma$ parameters following \cite{Hsu10apractical}, while for Random Forest we simply manually tried different sizes of decision forests.
\subsubsection{Other approaches}
The above approaches based on a ML classification algorithm have been compared against the approaches described in Section \ref{prelim}.

In order to reproduce the \cphydra\ approach, we computed the scheduling that it would have produced for every instance of the test set and simulated this schedule. Since this approach does not scale very well w.r.t. the size of the portfolio we were able to simulate this approach only for small portfolios (i.e. containing less than 9 solvers). To compute the PSI and AST we did not take into account the time needed to compute the schedule; therefore the results of \cphydra\ presented in this paper can be considered only an upper bound of its real performances.

We simulated the SATzilla approach 
by developing a MATLAB 
implementation of the cost-sensitive classification model described in \cite{DBLP:conf/sat/XuHHL12}, with the only 
exception that ties during solvers comparison are broken by selecting the solver that in general solves the 
largest number of instances.
We employed Mistral as a backup solver in case the solver selected by SATzilla ended prematurely.

To simulate the 3S approach we did not use the original code to compute the static schedule since it is not publicly available. To compute the schedule of solvers we used instead the mixed integer programming solver Gurobi~\cite{gurobi} to solve the problem described in \cite{DBLP:conf/cp/KadiogluMSSS11}. However, in order to reduce the search space, instead of using the column generation method as used by the developers of 3S, we imposed an additional constraint requiring every solver to be run for an integer number of seconds.
In this way it was possible to obtain a good enough schedule of solvers to run for 180 or fewer seconds.\footnote{Note that even 3S is using a column generation technique to reduce the size of the problem in spite of the optimality of the solution.} If the instance was not solved in this time window the solver that solved the majority of the most $k$-similar instances was used for the remaining time (possible ties were broken by minimizing the average solving time) and, in case of failures, Mistral was used as a backup solver.

Thanks to the code kindly provided by Yuri Malitsky, we were able to adapt ISAC 
cluster-based techniques to create a solver selector using the ``\emph{Pure Solver Portfolio}" approach as done for SAT problems in \cite{DBLP:conf/cpaior/MalitskyS12}. We clustered the training instances and we mapped every resulting cluster to the corresponding best solver. 
 For every test instance we determined the closest cluster according to the Euclidean distance and then we used the solver associated to it first. Also in this case Mistral was used as a backup solver in case of failures of the first solver.

\section{Results and Assessments}
\label{results}
This section presents the experimental results of our work.

\begin{figure}[t]
\centering
\begin{subfigure}{0.47\textwidth}
 \centering
 \includegraphics[width=\textwidth, clip, trim=40mm 70mm 40mm 70mm]{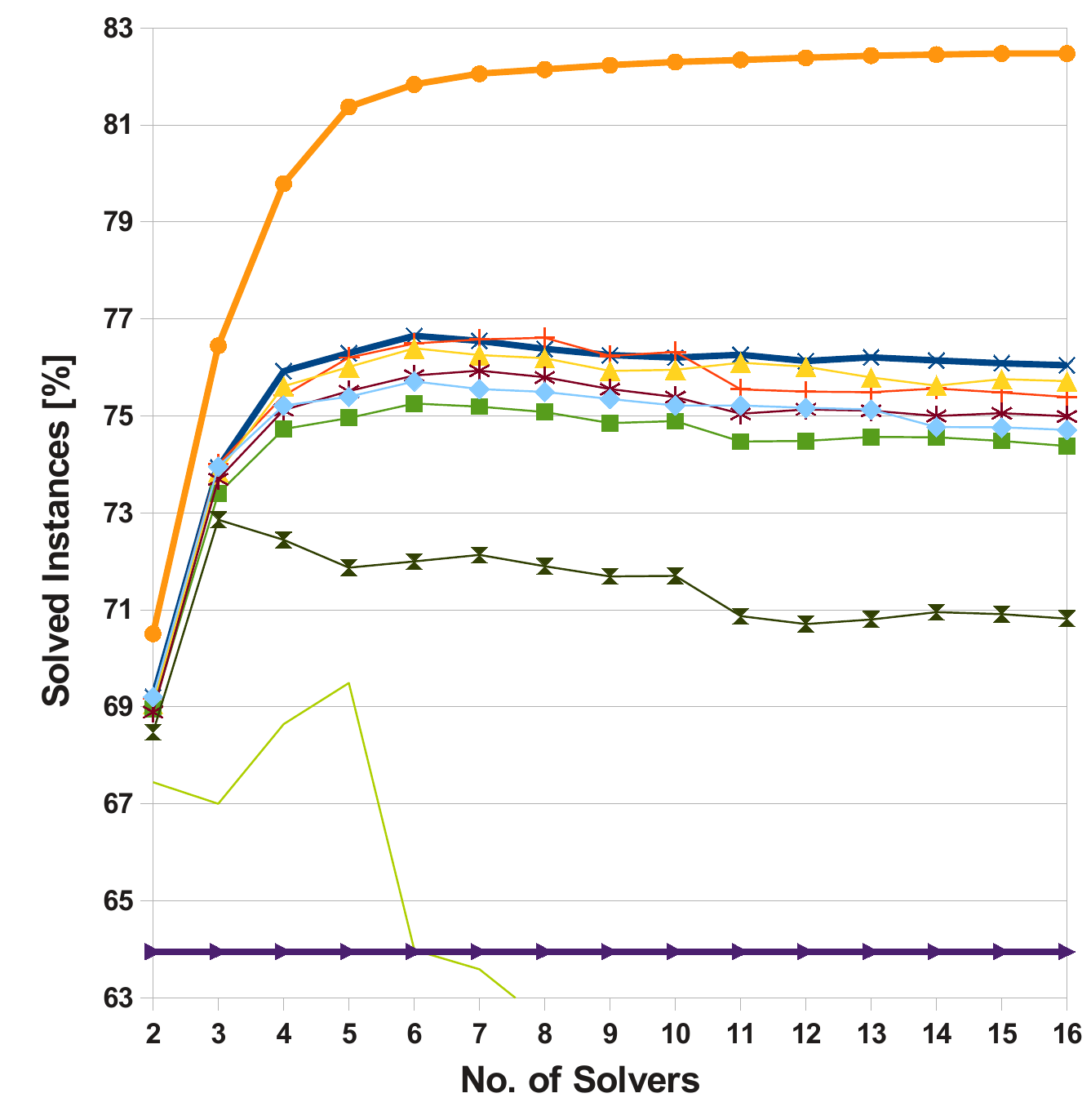}
 \caption{}
 \label{figure:PSI}
\end{subfigure}
\begin{subfigure}{0.47\textwidth}
 \centering
 \includegraphics[width=\textwidth, clip, trim=40mm 70mm 40mm 70mm]{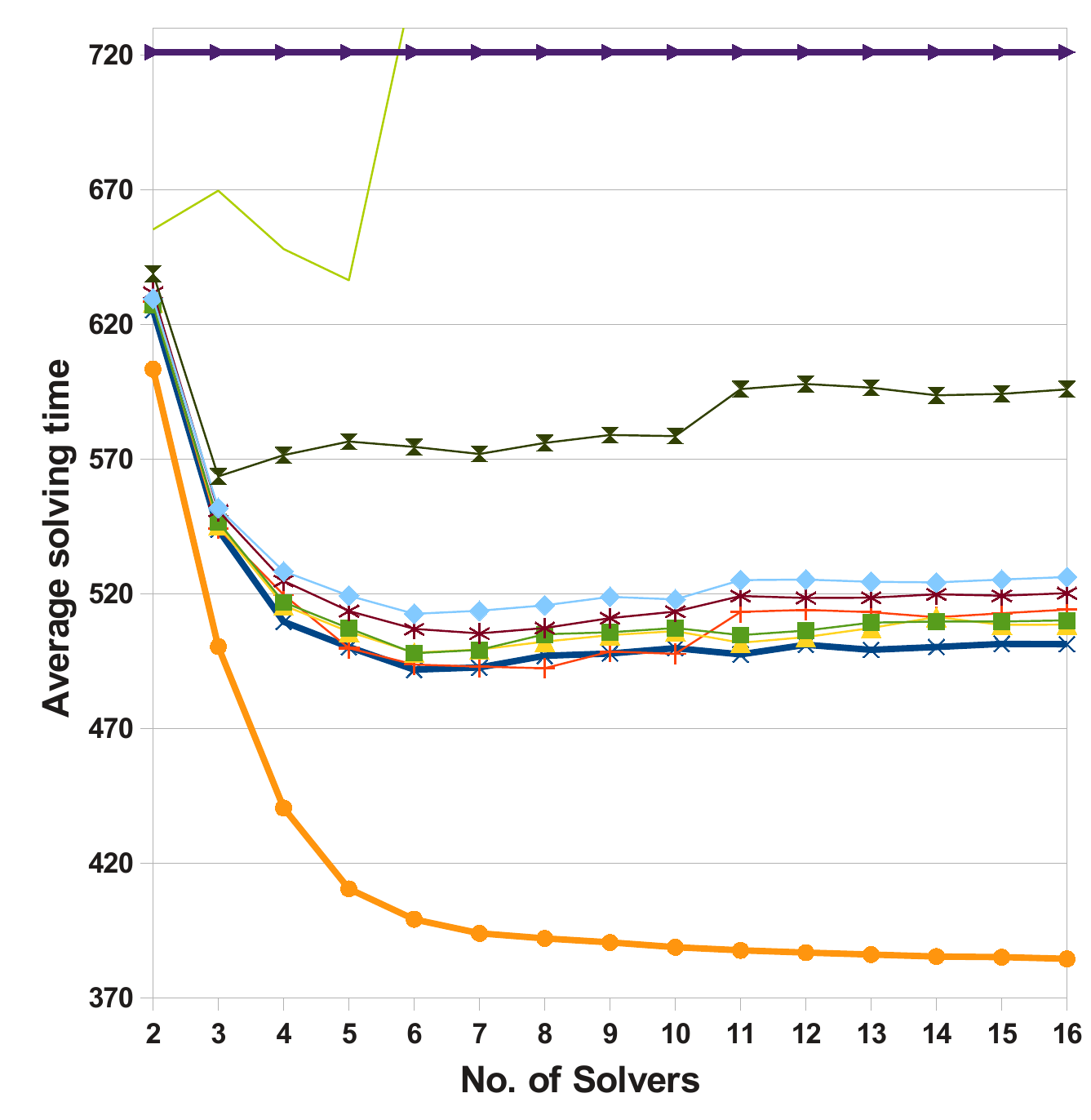}
 \caption{}
 \label{figure:AVG}
\end{subfigure}
\begin{subfigure}{0.48\textwidth}
 \centering
 \includegraphics[width=\textwidth, clip, trim=30mm 90mm 30mm 90mm]{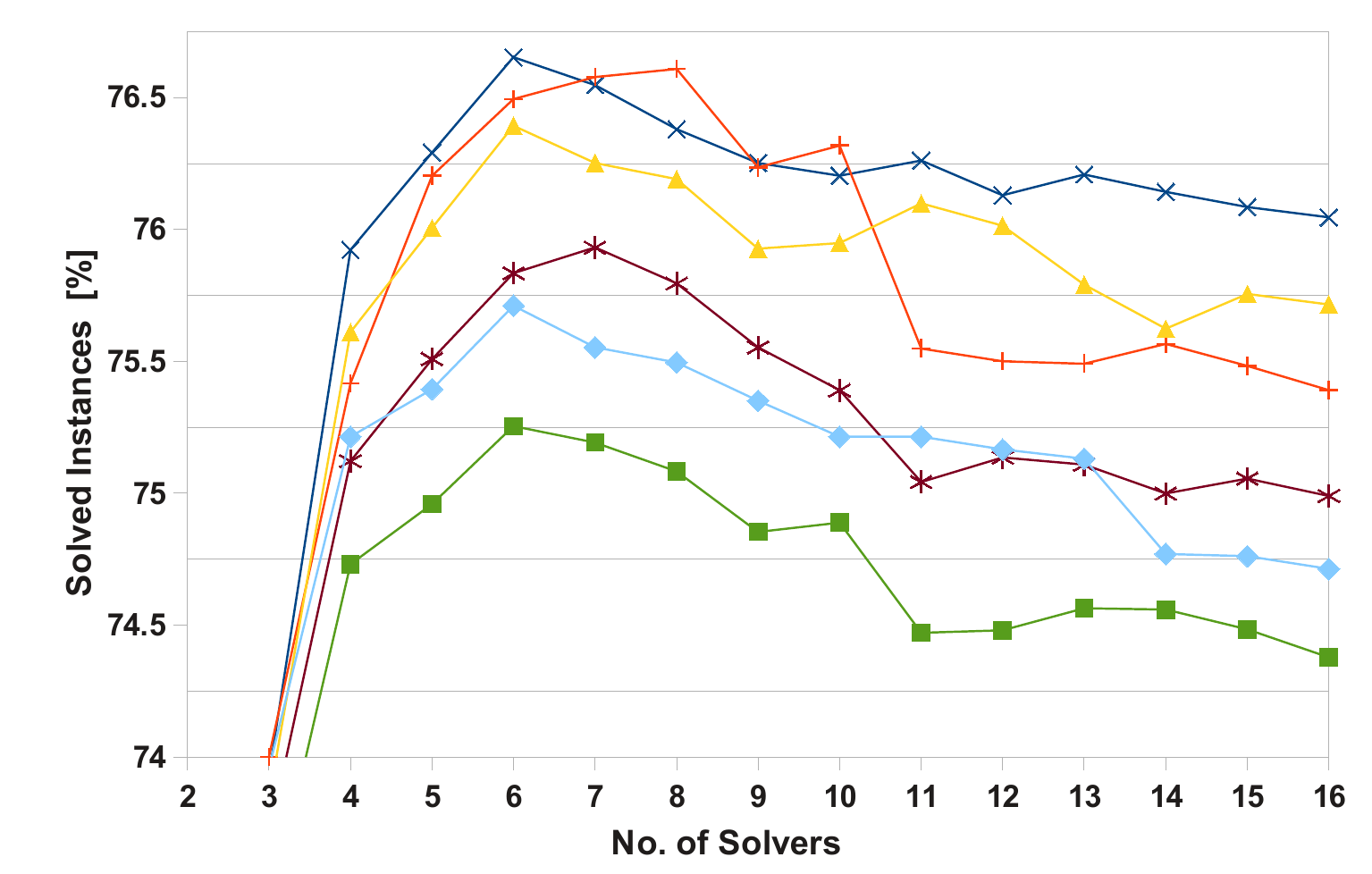}
 \caption{}
 \label{figure:PSI_detail}
\end{subfigure}
\begin{subfigure}{0.48\textwidth}
 \centering
 \includegraphics[width=\textwidth, clip, trim=30mm 90mm 30mm 90mm]{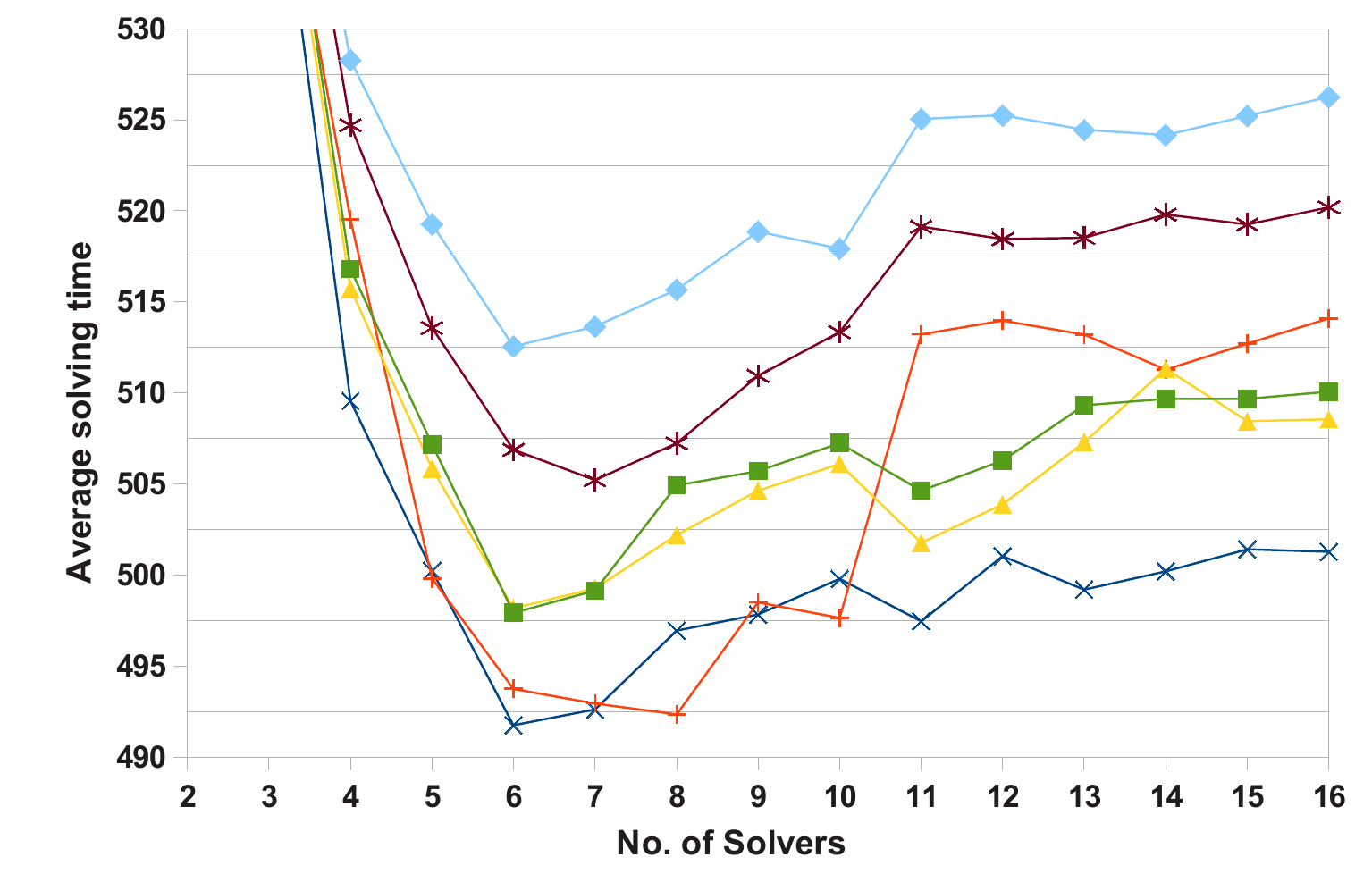}
 \caption{}
 \label{figure:AVG_detail}
\end{subfigure}
\begin{subfigure}{\textwidth}
 \centering
 \includegraphics[width=0.7\textwidth, clip, trim=30mm 125mm 30mm 125mm]{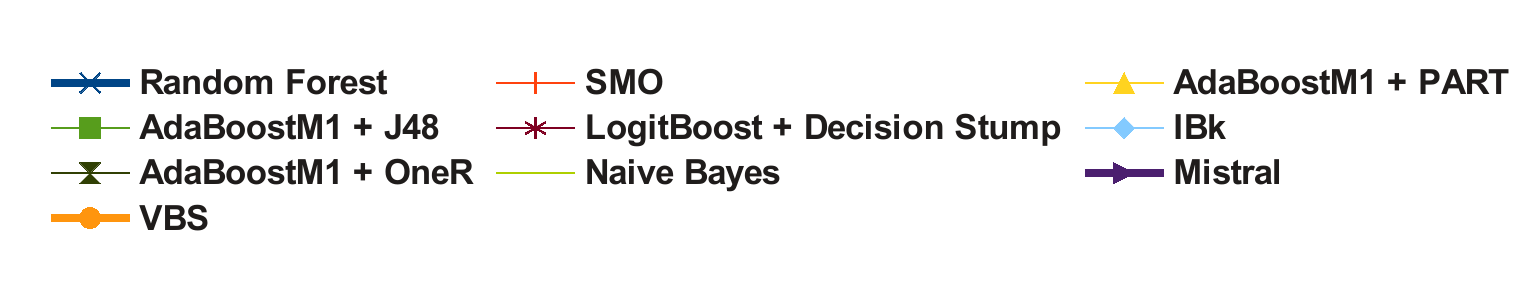}
\end{subfigure}
\caption{Performances of the portfolio approaches based on classifiers}
\end{figure}
Figures \ref{figure:PSI} and \ref{figure:AVG} show respectively the PSI and AST performances of the approaches using off-the-shelf classifiers,
setting as baselines the performances of Mistral with a time cap of 1800 seconds and of the Virtual Best Solver (VBS), i.e. an oracle that for every instance always chooses the best solver.\footnote{Note that, unlike all the other approaches, the VBS does not consider the time required to extract the instance features.
} We report the results of all the classifiers listed in the previous Section eventually boosted by a meta classifier whenever its use improved the performances.
Since there are some approaches that have similar performances and can not be distinguished easily, in Figures \ref{figure:PSI_detail} and \ref{figure:AVG_detail} we report these methods to allow a better comparison between their performances.


From these figures we can see that almost all the approaches outperform the simple use of Mistral, both in the number of instances solved and in the average solving time. The only exception is the Naive Bayes approach that for portfolios with more than 6 solvers makes so much errors that its performances became worse than the ones of Mistral.
The best approach was Random Forest with a portfolio of 6 solvers that solved $76.65\%$ of the instances (the VBS solved  $82.47\%$ of the instances with a portfolio of $14$ solvers or more).
However, other approaches which are built on top of support vector machines (SVMs) or meta classifiers have similar performances (e.g. the SMO approach using SVM is able to solve $76.61\%$ of the instances with a portfolio of $8$ solvers).

To be able to establish if an approach was better than another one in a statistically significant way we used the Student's paired $t$-test. Given a portfolio approach and a portfolio size, we considered all the $k = 22735$ instances of the 25 test sets and obtained the corresponding sample data as a binary distribution $x_1, \dots, x_k$ in which a variable $x_i$ is 1 if and only if the corresponding instance was solved by the portfolio solver in less than 1800 seconds. 
Then, given two different samples $X = \langle x_1, \dots, x_k \rangle$ and $Y = \langle y_1, \dots, y_k \rangle$ derived 
from two corresponding portfolio approaches and sizes, we considered them statistically significant if the $p$-value of the paired $t$-test on $X$ and $Y$ 
was below $0.05$.



Comparing the portfolio approaches by fixing the portfolio size we noticed that for portfolios of size ranging from 5 to 10 the performances of Random Forest are not statistically significant w.r.t. SMO, while they are significant if compared to all other approaches. This confirms the similarity of the performance of the Random Forest and SMO classifiers as can be seen from the plots.


As far as the portfolio size is concerned, we noticed that for every classifier the prediction becomes inaccurate after a given size, thus hindering the performance of the approach. So, even though the use of a larger portfolio means that potentially more instances could be solved, the best performances were obtained by using portfolios from 6 to 8 solvers. For some classifiers the drop of performances was quite significant: for instance the SMO classifier with a portfolio of 16 solvers solved $1.22\%$ less instances than the version with a portfolio of 8 solvers.
We also compared the peak performances (in terms of PSI) obtained by varying the portfolio size for a fixed portfolio approach. 
It turned out that for Random Forest the difference in the peak performance (obtained with 6 solvers) 
was not statically significant only w.r.t. the case of  7 solvers, 
while for all other pairs (6, k) with $k\neq 7$ such a difference was statically significant. In a similar way, for SMO only the difference between 
the case of 8 solvers (peak performance) and 7 solvers was not statically significant, while in all other cases it was. These results confirm from a  statistical point of view the intuition that it is sometimes better to chose a smaller portfolio than a larger one.

Since our dataset was imbalanced (i.e. the class of instances where Mistral was the best solver was far greater than the class of instances of other solvers) we have tried to apply oversampling techniques (Synthetic Minority Over-sampling Technique \cite{DBLP:journals/corr/abs-1106-1813} to be precise) to boost the accuracy of the classifiers. This however has not led to improvements in the metrics we used to evaluate the performances of the classifiers.

\begin{figure}[t]
\centering
\begin{subfigure}{0.48\textwidth}
 \centering
 \includegraphics[width=\textwidth, clip, trim=40mm 85mm 40mm 85mm]{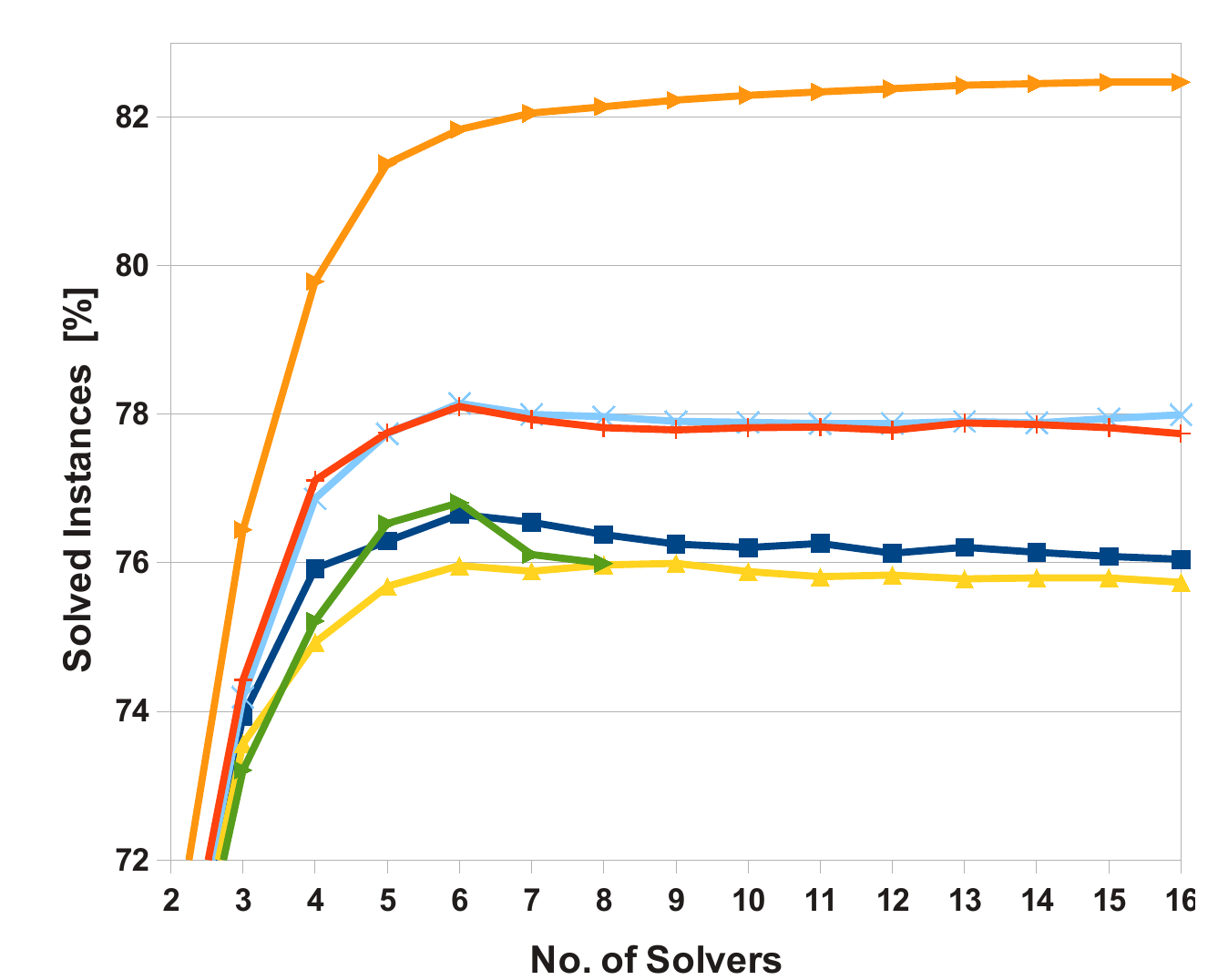}
\end{subfigure}
\begin{subfigure}{0.48\textwidth}
 \centering
 \includegraphics[width=\textwidth, clip, trim=40mm 85mm 40mm 85mm]{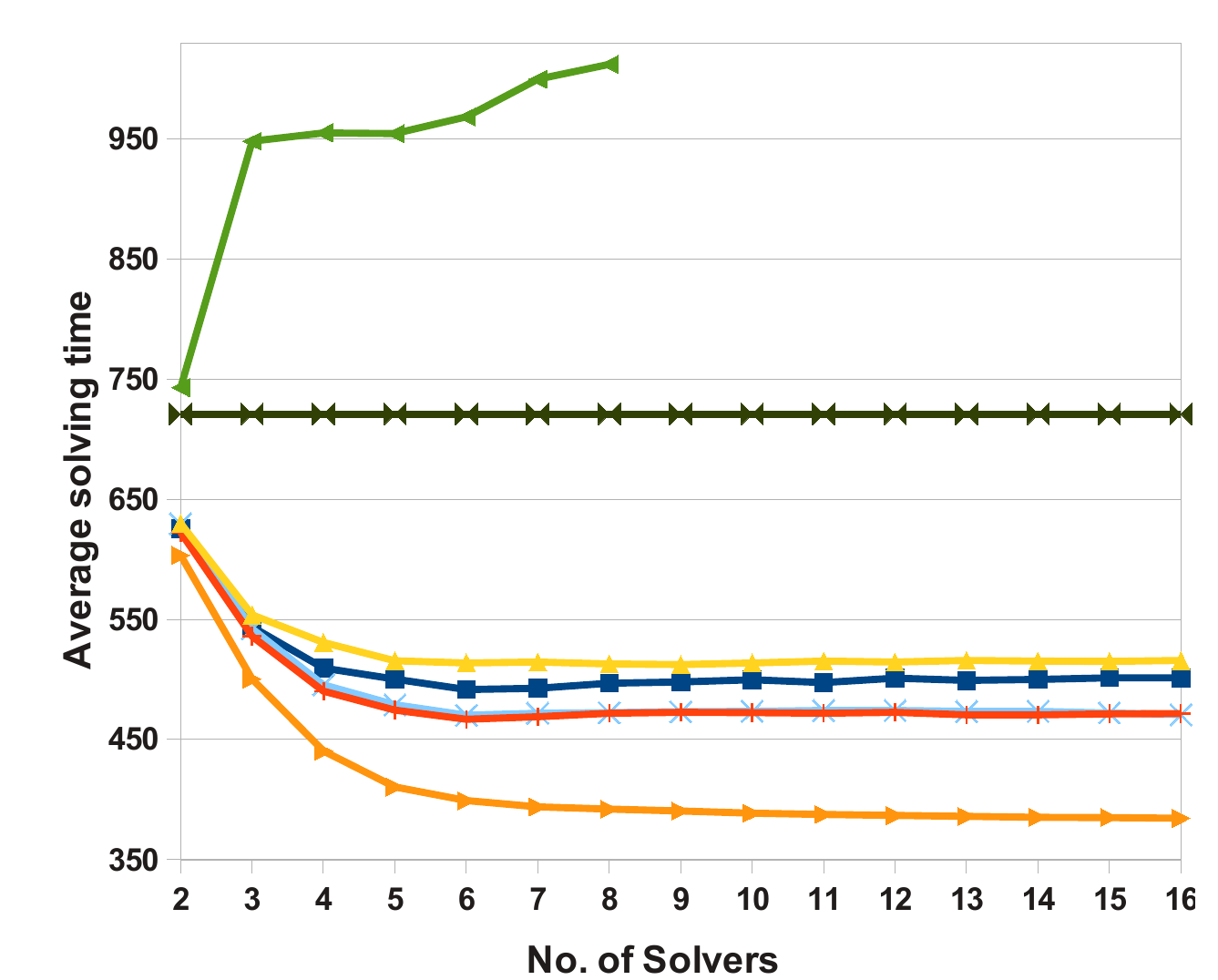}
\end{subfigure}
\begin{subfigure}{\textwidth}
 \centering
 \includegraphics[width=0.5\textwidth, clip, trim=50mm 130mm 50mm 130mm]{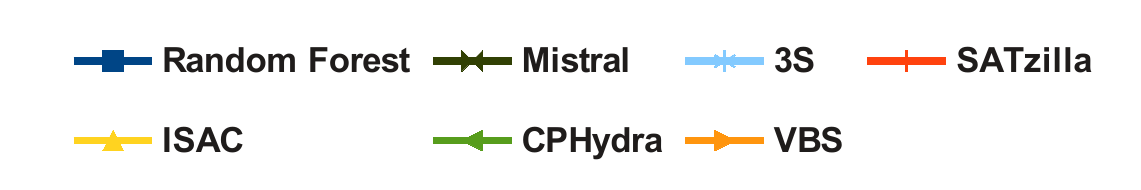}
\end{subfigure}
\caption{Performances of portfolio approaches}
\label{figure:others}

\end{figure}

In Fig. \ref{figure:others} we show the comparison between the approaches of SATzilla, ISAC, 3S, \cphydra\ and the approach which used Random Forest as solver selector. As already stated, due to the computational cost of computing the schedule of solvers, for \cphydra\ we report the results obtained using just less than $9$ solvers.\footnote{For the case of 8 solvers the computation of the schedule did not terminate in 24 hours for 4 instances that consequently were not considered for evaluating the \cphydra\ performances.}


In this case it is possible to notice that the best approaches used in SAT, namely 3S and SatZilla, have peak performances. 3S is able to solve usually few more instances than SatZilla (3S have a peak PSI of $78.15\%$ against the $78.1\%$ peak performance of SatZilla) while SatZilla is usually faster (the AST of Satzilla with a portfolio of size 6 was $466.82$ seconds against the $470.30$ seconds of 3S). Even though conceptually 3S and SatZilla are really different they have surprisingly close performances. This is confirmed also from a statistical point of view since their performances are not statistically significant if they use the same portfolio.
3S and SatZilla are instead statistically better than all the other tested approaches for portfolios of size greater than 3 (3S is able to close $26\%$ of the gap of Random Forest w.r.t. the VBS). Moreover, the decay of performances due to the increase of the portfolio size is less pronounced that what usually happens when a classifier is used as a solver selector. As in the classification based approaches, the peak performance was reached with a relatively small portfolio (6 solvers) and the peak performances of both 3S and Satzilla are statistically significant w.r.t their performances with different portfolios sizes.
The performances of ISAC are slightly worse than those of Random Forest: the maximum PSI reached was $75.99\%$ while the Random Forest approach obtained $76.65\%$.

As far as \cphydra\ is concerned we saw that it solved the maximum number of instances with a portfolio of size $6$ reaching a PSI of $76.81\%$ that was slightly better than the peak performance obtained by Random Forest and SMO, even though not in a statistically significant way. After reaching the maximal number of solved instances \cphydra\ performances are decreasing and in a real scenario they would be rather poor since computing the optimal solvers schedule can consume a lot of time.
From Figure \ref{figure:others} it is possible to note that \cphydra\ differs from other approaches because it is not developed to minimize the average solving time. There is no heuristic to decide which solver needs to be run first in order to minimize the solving time.
For this reason, \cphydra\ is the only approach,  among those we have considered, where the PSI and AST values have a positive correlation.
Indeed, the Pearson correlation coefficient between PSI and AST values 
is $0.921$, which means that PSI and AST are almost in linear relationship.
Conversely for the other best performing approaches 
the correlation coefficient was always below $-0.985$ meaning that minimizing the average solving time was like requiring to maximize the number of instances solved and vice versa.

The considerable performances achieved by 3S and SatZilla encouraged us
to combine both approaches using the fixed-time static solver schedule of 3S with the dynamic selection 
of a long-running solver made using SatZilla approach. To our surprise, the performances of this combined
approach did not improve the individual performances of 3S and SatZilla.



Even though the goal of this paper was just to compare the performances of different portfolio approaches, we would like to spend also few words on the time needed to build a prediction model. Obviously, if this task can be performed offline, maybe using cluster to parallelize the workload, the time needed to build a model is not very significant. This however could change in a more dynamic scenario like the case of a system that at run time exploits the instances it previously solved to increase its efficiency and performances.
The approaches studied in this paper can be adapted to this dynamic scenario by updating the models periodically. Clearly, if the update of a model requires hours the system may not be very responsive,  therefore portfolios approaches where the time to build a model is short are preferable in dynamic scenarios.


\begin{figure}[t]
\centering
\includegraphics[width=0.95\textwidth, clip, trim=15mm 90mm 15mm 90mm]{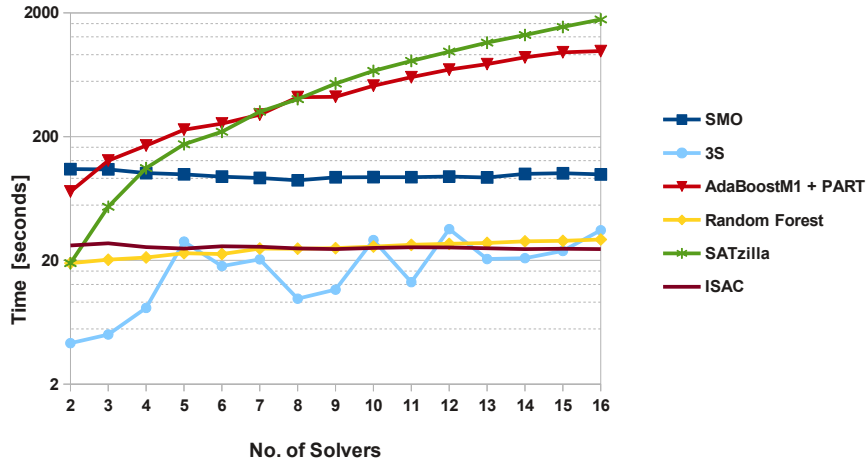}
\caption{Times taken by portfolios approaches to create one prediction model}
\label{figure:times}
\end{figure}

In Figure \ref{figure:times} we report the times needed by different approaches to build the prediction model of one fold of the dataset.
Without considering \cphydra\, where the model is computed for every instance solving an NP-hard problem, among the tested methods the one that employs the longest time to build a model is SATzilla. Even though this task can be easily parallelizable,  the computation for a portfolio with 16 solvers using only one machine required more than an hour.\footnote{Note that we used a MATLAB implementation of Weighted Random Forest: the SATzilla original code may be more efficient.} As can be also seen from the plot (please note the logarithm scale of the y-axis) the building time grows (quadratically) w.r.t. the portfolio size since SATzilla needs to compute for every pair of solvers a weighted random forest of trees. The same correlation between model building times and portfolio sizes happens also for the AdaBoostM1 classifier while instead for the other approaches the cost of building models does not depend on the portfolio size.
As far as the 3S approach is concerned, in Figure \ref{figure:times} we present the times needed to compute the static schedule for one fold. As already stated this is an NP-hard problem and its solution times may heavily depend on the specific training data having an erratic and unpredictable behavior. 
So even though SATzilla and 3S were clearly the winners of the comparison between the different portfolio approaches for CSPs, it could be the case that for dynamic scenarios other approaches like Random Forest or online machine learning classification algorithms \cite{onlineML} could be more useful.

\section{Related work}
\label{related}
For the sake of clarity in Section \ref{prelim} we presented only the portfolio approaches we used. In this section we will briefly present other relevant works on portfolios.

Gebruers et al.~\cite{Gebruers-Et-Al-2005} use case-based reasoning to select solution strategies for CSPs. They consider a portfolio of 12 strategies and apply a $k$-nearest neighbor algorithm to predict the strategy that should speed up the search of solutions for the social golfer problems.
Streeter et al.~\cite{DBLP:conf/aaai/StreeterGS07} use instead optimization techniques to produce a 
schedule of solvers that should be executed in a specific order, for specific amounts of time, 
in order to maximize the probability of solving the given instance.
In \cite{DBLP:conf/ecai/GuerriM04}, a classification-based algorithm selection for a specific CSP 
is studied. Given an instance of the \emph{Bid Evaluation Problem} (BEP), the objective is to be able to 
decide a-priori whether an Integer Programming (IP) solver, or a hybrid one between IP and CP 
will be the best. Such a selection is done on the basis of the instance structure which is determined 
via (a subset of) 25 static features derived from the constraint graph~\cite{DBLP:conf/cp/Leyton-BrownNS02}. 
These features are extracted on a set of training instances and the corresponding best approach is identified. 
The resulting data are then passed to a classification algorithm that builds decision trees.
An alternative model-based portfolio approach presented in~\cite{silverthorn:aaai2010} addresses the problem 
of predicting the solver performances on a given instance using a \emph{Dirichlet Compound Multinomial} (DCM) 
distribution to create a schedule of solvers (for such an instance).


\section{Conclusions}
\label{concl}

In this work we have implemented different portfolio approaches for solving 
Constraint Satisfaction Problems (CSPs). These approaches have been obtained both by 
using machine learning techniques and adapting to CSPs other algorithms proposed in 
the literature, mainly in the SAT solving field.
We have evaluated and compared the different approaches by considering a dataset 
consisting of 4547 instances taken from
two different kind of constraint competitions and a selection of 22 versions of different solvers.
The portfolio approaches were evaluated on the 
basis of the number of pro\-blems solved and the time taken to solve them.
The experimental results show that the approaches that won the last two SAT competitions, namely SATzilla and 3S, are the best ones among those considered in this paper, both for the instances solved and the time needed to solve them. However approaches using off-the-shelf classifiers as solver selector are not that far from the best performances and can potentially be used in scenarios were the time needed to build the model to make the predictions matters.
Another interesting empirical fact is that, for all but one the portfolio approaches considered here, there was a strong anti-correlation between the average solving time and the number of solved instances. Minimizing the average solving time in this setting can therefore lead to solve more instances and vice versa.

We are aware of the fact that our results are not as exhaustive as those existing in the SAT field. Indeed,  the number of solvers that we used is relatively small. 
Moreover our solvers are not so different. 
Also the number of CSP instances that we used to evaluate the portfolio approaches is smaller that the thousands of problems that are available 
in the SAT community. Thus our results have a more limited significance than the results existing for SAT approaches. 
However we believe that we made a first step towards a clarification of the importance of the portfolio approaches for solving 
CSPs.  As a future work we plan to extend the number of protfolio approaches by considering also the dynamic schedule approach of 3S \cite{DBLP:conf/cp/KadiogluMSSS11}, the regression based approach of the previous version of SATzilla  and other approaches which are not based on feature extraction like~\cite{silverthorn:aaai2010}.
Moreover we are also interested in studying the impact of instance-specific algorithm configuration tools like ISAC or HYDRA~\cite{DBLP:conf/aaai/XuHL10} in the CSP field by allowing the automatic tuning of search and other solver parameters to boost the solver performances.


\bibliographystyle{plain}
\bibliography{biblio}

\end{document}